# Numeral Understanding in Financial Tweets for Fine-grained Crowd-based Forecasting

Chung-Chi Chen,[1] Hen-Hsen Huang,[1] Yow-Ting Shiue,[1] Hsin-Hsi Chen[1,2]
[1]Department of Computer Science and Information Engineering,
National Taiwan University, Taipei, Taiwan
[2]MOST Joint Research Center for AI Technology and All Vista Health care, Taiwan
{cjchen, hhhuang}@nlg.csie.ntu.edu.tw, orina1123@gmail.com, hhchen@ntu.edu.tw

***Abstract*—Numerals that contain much information in financial documents are crucial for financial decision making. They play different roles in financial analysis processes. This paper is aimed at understanding the meanings of numerals in financial tweets for fine-grained crowd-based forecasting. We propose a taxonomy that classifies the numerals in financial tweets into 7 categories, and further extend some of these categories into several subcategories. Neural network-based models with word and character-level encoders are proposed for 7-way classification and 17-way classification. We perform backtest to confirm the effectiveness of the numeric opinions made by the crowd. This work is the first attempt to understand numerals in financial social media data, and we provide the first comparison of fine-grained opinion of individual investors and analysts based on their forecast price. The numeral corpus used in our experiments, called FinNum 1.0[1], is available for research purposes.**

## I. INTRODUCTION

Text-based FinTech (financial technology) is a hot topic that attracts much attention in recent years. Many related workshops and shared tasks co-located with influential conferences, including SemEval 2017 Task 5 (co-located with ACL'17), FNP 2018 (co-located with LREC'18), FiQA 2018 (co-located with WWW'18), ECONLP 2018 (co-located with ACL'18), and FinNum 2019 (co-located with NTCIR-14) showing a great potential on applying natural language processing (NLP) technologies to the financial domain.

Social trading, a process that helps online investors analyze financial instruments and make decisions, is a key financial service [1]. In the social trading platform, investors discuss their trades, share their trading strategies, and provide their opinions on certain financial instruments such as stocks, bonds and foreign exchanges. With social trading process, investors get more information than before.

Numerals play important roles in financial analysis processes such as determining the intrinsic values of financial instruments and forecasting the movement of asset prices based on the past market data. For example, to measure the value of a stock, investors employ information from company's financial statements. To evaluate the bond, investors may concern with the macroeconomic data like interest rate. To predict the price trend, investors may use technical indicators calculated with history price or analyze price charts and look for the embedded patterns. In this paper, we attempt to leverage the numeric opinions made by the crowd by understanding the meanings of numerals.

As the idea of crowdsourcing, if we could catch on to the view of most investors about certain financial instrument, we will have lower uncertainty and more confidence when predicting the price movement of this instrument. For the purpose of capturing the crowd opinions, financial tweets are adopted as the source of investors' views. In this paper, we classify the numerals in financial tweets into 7 categories, including Monetary, Percentage, Option, Indicator, Temporal, Quantity, and Product/Version Number. We further extend 4 of these categories to several finer classes. Based on the taxonomy, we aim to disambiguate the meanings of numerals in financial tweets. A variety of neural network models are explored for the two classification tasks.

Although numerals are important in financial documents, few of the previous works make attempt to understand the numerals in financial text. In financial statements, numerals are often provided in a structural form that is easier to sort out the information. By contrast, numerals in social media data are unstructured and noise-induced. To capture the investors' opinions and understand the numerals at semantic level, fine-grained classification of numerals is indispensable. Consider the numerals in the following tweet (T1) as an example:

**(T1)** *$TSLA 256 Break-out thru 50 & 200- DMA (197-230) upper head res (274-279) Short squeeze in progress Nr term obj: 310 Stop loss:239.*

where 256, 197, 230, 274, 279, 310 and 239 represent the prices of TSLA, but they stand for different meanings. Numeral 256 is the quote of TSLA. Numerals 197 and 230 are the values of Different of Moving Average (DMA), one of the technical indicators, with parameters 50 and 200 respectively. Numerals 274-279 is the range of resistance. That is, this tweet writer thinks that if the price increases up to this range, the upward trend will be impeded. Numeral 310 is the near-term objective of the TSLA price. Numeral 239 is the stop loss price, which means that this tweet writer will sell out the position of TSLA once the price goes downward below 239. In this 25-word tweet, the nine numerals can be separated into 2 categories, price and parameter of technical indicator. Furthermore, the price could be further classified into five subcategories, including quote, value of technical indicator, resistance range, forecast price, and stop loss price.

Unlike name entity recognition (NER) , which targets all kinds of name entities, we focus on the numerals in financial social media data. Sekine [2] extended the taxonomy of named entity with attribute information, and included 3 kinds of numeral expressions in the financial domain (i.e. money, stock index, and percent). However, their taxonomy can hardly capture the opinion of investors. For example, 310 and 239 in (T1) are classified into stock index in their taxonomy. In our taxonomy, by contrast, 310 is annotated as "forecast" price, and 239 is "stop loss", denoting the investor's prediction of the stock price and the bottom level used to limit loss, respectively. Such a distinguishing on the subjective analysis results of individual investors not only provides crucial information for decision making, but also enables us to construct a trading model.

[1] A numeral corpus, called FinNum 2.0, is released by the following URL: http://nlg.csie.ntu.edu.tw/nlpresource/FinNum/. The performance of our approach by using FinNum 2.0 is also reported.

To the best of our knowledge, this paper is the first paper focusing on understanding the meaning of numerals in financial social media data. Our contributions are four-fold. Firstly, we propose a numeral taxonomy for fine-grained opinion mining on financial social media data. Secondly, we annotate the numerals in financial tweets with this taxonomy and construct a dataset for experiments. Thirdly, we conduct comprehensive experiments to compare the performance of different classification models in coarse-grained and fine-grained tasks. Lastly, with the proposed dataset, FinNum, and methods, automatically evaluating the fine-grained opinions of individual investors become possible, and the first empirical study results are demonstrated in this paper.

This paper is organized as follows. Section II investigates related works. Section III provides the tailor-made taxonomy for numerals in financial domain. Section IV describes the details of the annotated dataset. Section V shows the experimental methods. Section VI records the experimental results. Analysis results of the experiments and the extended experiments are discussed in Section VII. In Section VIII, we evaluate the trading strategies based on crowd opinion. Section VIIII concludes this work.

## II. Related work

Temporal, a category of numerals, is one of the foci in previous work. Ling and Weld [3] propose an extractor for temporal information with probabilistic inference. Tourille et al. [4] attempt to extract numeral information from clinical documents. Davidov and Rappoport [5] extract numerical information like size and depth from the web and experiment on the question answering task. Madaan et al. [6] deal with numeral relation extraction, and propose the state-of-the-art systems to extract the geopolitical relations between numeral and country. However, none of the previous work investigated the meaning of the numerals in financial domain as intensive as we do. Murakami et al. [7] use stock prices to generate market comments. That shows the importance of numerical information in finance.

Sentiment analysis, a widely-studied topic in the NLP community, is one of the applications of numeral understanding. Bollen et al. [8] show that the public mood on Twitter is correlated to Dow Jones Industrial Average value. Li et al. [9] introduce sentiment of news articles into their model, and indicate that sentiment information do help the accuracy of predicting stock price. Khedr et al. [10] use news sentiment analysis results to predict the behavior of the stock market. Although there are several research about sentiment analysis of financial data, none of them do the in depth work to capture the fine-grained opinions of investors.

This paper is the first attempt to understand numerals in financial social media data. It is expected to be useful for fine-grained opinion analysis. Because numerals contain crucial information in financial documents, they are quite important when analyzing the financial instruments. With the classified numerals, we can obtain more opinion information from investors, and use these opinions to do further research. The application scenarios are provided in Section VIII.

## III. Numeral Taxonomy

For the in depth mining of investors' views, the first challenge is the implications of numerals. We sort out the numerals in financial tweets by expert's experience. A total of 7 categories are proposed for numerals, and four of them are further extended to various subcategories, as shown in Table I. The most important category for financial tweets is Monetary, which is divided into 8 subcategories. The details are elaborated in subsequent sections.

TABLE I. NUMERAL DISTRIBUTION IN OUR DATASET

| Category | Subcategory | Number of instances | % |
|---|---|---|---|
| Monetary | money | 77 | 5.74 |
| | quote | 140 | 10.44 |
| | change | 49 | 3.65 |
| | buy price | 68 | 5.07 |
| | sell price | 34 | 2.54 |
| | forecast | 122 | 9.1 |
| | stop loss | 12 | 0.88 |
| | support or resistance | 107 | 7.98 |
| Percentage | relative | 192 | 14.32 |
| | absolute | 86 | 6.41 |
| Option | exercise price | 45 | 3.36 |
| | maturity date | 8 | 0.6 |
| Indicator | | 48 | 3.58 |
| Temporal | date | 197 | 14.69 |
| | time | 39 | 2.91 |
| Quantity | | 100 | 7.46 |
| Product | | 17 | 1.27 |

### A. Monetary

The Monetary category contains the following 8 subcategories: "money", "quote", "change", "buy price", "sell price", "forecast", "stop loss" and "support or resistance". The ideas to distinguish these subcategories are: (1) "money", "quote" and "change" just describe a status; (2) the other subcategories present the opinions of a tweet writer. As our discussion for the tweet (T1), 256, 197 and 230 just quote the price of TSLA and the DMA. By contrast, 274, 279, 310 and 239 are the analysis result of the writer. Moreover, these numerals have different meanings. Numerals 274 and 279 are annotated as "support or resistance", 310 belongs to "forecast", and 239 is classified as "stop loss". In tweet (T2), both 800 and 1 are annotated with "money" label. In tweet (T3), "+2.00" is an example for subcategory "change".

**(T2)** *MarketWatch: RT wmwitkowski: Guess who sold off about $800 million in $MDLZ after losing about $1 billion on $VRX???*

**(T3)** *$NVDA Sunday watchlist entry from MON +2.00 and going now someone bought lots of calls MON*

The identification of "buy price" and "sell price" can help us understand the performance of the writer. Based on the performance information, we can give different weights for the opinion of each investor. 137.89 in (T4) is an instance for "buying" subcategory. 36.50 in (T5) is an example for "selling" aspect.

**(T4)** *$SPY Long 1/2 position 137.89*

**(T5)** *$KOG Took a small position hopefully a better outcome than getting kneecapped by $BEXP selling itself dirt cheap at 36.50*

Some investors "forecast" the price of the instruments depending on their analysis results. The numeral about the prediction of monetary will be classified into "forecast" subcategory. 14.35 in (T6) is an example for "forecast" subcategory. On the other hand, "stop loss" price is the price level that investors may close their positions.

**(T6)** *$CIEN, CIEN seems to have broken out of a major horizontal resistance. Targets $14.35.*

The concepts of support and resistance are always discussed in technical analysis. Merging these two terms into one subcategory is applicable because the investors who use technical analysis believe that when the price breaks up (down) the resistance (support), the resistant (supporting) price will become the support (resistance). This subcategory can help us indicate the boundary of price movement. 46 in (T7) is an instance for "support or resistance" subcategory.

**(T7)** *$CTRP, $46 Breakout Should be Confirmed with Wm%R Stochastic Up*

*B. Percentage*

There are many numerals about ratio in financial documents. For example, there are a lot of accounting ratio like P/E ratio, current ratio, and so on. All numerals about percentage information will be classified into Percentage category, and further extend into two subcategories, "absolute" and "relative". The numeral that indicates the proportion of a certain amount is classified into "absolute", while the numeral that stands for the change relative to original amount is classified into "relative". In tweet (T8), 10% and 7.5% are annotated as "relative", and 23% stands for "absolute".

**(T8)** *¢Den up almost 10% since Q1 and £áuro up around 7.5%, much more $ for $AAPL pocket. Remember 23% of Apple revenues comes from this two @jimcramer*

*C. Option*

Option is a popular instrument frequently discussed. According to the 2016 annual report of Chicago Board Option Exchange, the largest U.S. options exchanges, the annual trading volume of options in 2016 is up to 1 billion. Call and put are two most common options. Call (put) provides the right to buy (sell) the underlying asset at exercise price before maturity date. To capture the implications of investors' opinions, we propose two subcategories for Option category, "exercise price" and "maturity date".

Numeral 44 in tweet (T9) is the "exercise price" of the XLU April call, a kind of option, but not the quote for XLU. Thus, it is proper to annotate it independently. Assume the quote for XLU is 42 now. "Maturity date" is distinguished from ordinary "date" under the Temporal category because it implies more information of the investors. Although there are many strategies to trade options, the buyer of 46 call could be seen as having more confidence for the upward trend of XLU than the buyer of 44 call. APR.22 in tweet (T10) is the "maturity date" of the MSFT calls. The buyer of the May 22 calls may have a longer-term view on MSFT than the buyer of the APR.22 calls has. Capturing both information can also help us evaluate the performance of investors as "forecast" price in Monetary category.

**(T9)** *$XLU long April $44 calls*

**(T10)** *$MSFT those APR.22 CALLS were getting hot...*

*D. Indicator*

This category captures the parameters of the technical indicators. Popular indicators include moving average (MA), moving average convergence divergence (MACD) and relative strength index (RSI). Investors using technical analysis always mention technical analysis indicators in tweets to share their strategies or their analysis results. Different investors may use dissimilar parameters for the same indicator. In order to capture the price most investors pay attention to, we should identify the parameters being used. For example, (T1) has quoted 197 and 230 as the value of 50 and 200 DMA, while (T11) just remarks 5dma, 13dma, and 20dma. Thus, sorting out the parameters being used can help calculate the values of technical analysis indicators.

**(T11)** *$ATHX riding 5dma higher, dropping to 13dma at the dips, sign of a healthy advancing stock that stays above 20dma*

*E. Temporal*

Temporal information is also important in financial domain. The day most investor focusing on is the one with high volatility. For example, the day releasing earning information or the day announcing economics data. Thus, to capture the temporal information could help us capture the important date and time that many investors focus on. Many researches focus on temporal identification. Based on Timex3 [11], we classify Temporal category into two subcategories, "date" and "time". 8/17 in the tweet (T12) is "date", and 2 in the tweet (T13) is "time".

**(T12)** *$AAPL 8/17 gap filled*

**(T13)** *$AAPL wants lower. up waves getting smaller on the 2 min*

*F. Quantity*

Quantity information can help us know the position of an investor, and we can give the large weighting to the opinions held by persons who have large positions. Furthermore, the amount of sales is also the important information in accounting. For instance, the impact of selling 5,838 shares in the tweet (T14) may be more than that of just selling 10.

**(T14)** *$NTRS Insider Trading: Clair St Unloaded 5,838 Shares of Northern Trust Corporation (NASDAQ:NTRS)*

*G. Product/ Version Number*

The version of products may contain numerals. We can use the product information to compare importance of different tweets. For example, the tweet (T15), which discusses iPhone 7, may be more important than the tweet that discusses iPhone 4. This is because the news for the latest product always has a larger influence on the stock price. Thus, capturing the Product/ Version Number is one of the important tasks in understanding the topic discussed.

**(T15)** *If the camera is protruding like that, $AAPL is losing to #samsung #iphone7 #samsunggalaxy*

## IV. Data Annotation

We extract 707 unique tweets containing numerals from the dataset of SemEval-2017 Task5 [12], which is collected from Twitter and StockTwits (a popular social media platform for investors to share their ideas and strategies). The tweets selected by SemEval-2017 Task5 must contain at least one cashtag such as "$AAPL", which stands for the company Apple Inc. Totally, 1,341 numerals are annotated in the proposed dataset, FinNum 1.0.

### A. Inter Annotator Agreement

The dataset in this paper is annotated by three experts with financial domain knowledge. Totally 60.8% of numerals get consistent annotation result (i.e., all three annotators' decisions are the same), 32.9% of numerals get majority of annotation result (i.e., two annotators' decisions are the same), and only 6.3% of numerals have inconsistent annotation result (i.e., three annotators' decisions are different).

The Kappa agreement between each two annotators are 70.30%, 69.75% and 67.07%. It is considered as substantial agreement [13]. To deal with the inconsistent numerals, three experts discuss one by one and select the final annotation from the three annotation decisions of their original choices.

The subcategories in Monetary category are the hardest to assign, especially, between the "quote" subcategory and other subcategories. This result indicates that the numeral containing which kinds of the opinion of investors sometimes may be a discussible question. $425 in (T16) shows an example that assigned as "quote", "forecast", and "support and resistance" in first round of annotation. After discussed by experts in second round, it is annotated as "support and resistance" subcategory.

**(T16)** *$AAPL could be testing $425 in days*

### B. Annotation Results

Table I shows the distribution of each category and their subcategories. As we discussed in Section III, Monetary category is the most important category in financial tweet. It occupies 45.4% of the annotated numerals. Percentage and Temporal account for 20.73% and 17.6%, respectively. Total 3.96% and 3.58% of numerals are annotated as Option and Indicator, two special categories for financial tweets.

TABLE II. Keyywords for Feature

| | |
|---|---|
| **Key_p** | %, percent, pc |
| **Key_r** | up, down , decline, increase, growth, gain, lose, +, - |
| **Key_m** | january, jan, february, feb, march, mar, april, apr, may, june, jun, july, jul, august, aug, september, sep, october, oct, november, nov, december, dec |
| **Key_i** | ma, dma, sma, ema, rsi, ichimoku … |
| **Key_d** | day, week, month, year, mos, yrs |
| **Key_t** | second, minute, hour, p.m., a.m. |

## V. Methods

This section shows the models for numeral classification based on the contextual information of the target numeral. We introduce some features for numerals classification, the data used to train word vectors, and the support vector machine (SVM), convolutional neural network (CNN), and recurrent neural network (RNN) classifiers.

### A. Features

Some features for certain categories and their subcategories will be introduced in this subsection. We convert all characters in tweets to lowercase when extracting features.

#### 1) Features for Percentage

To present the percentage, tweet writers often add some keywords or symbols following a numeral. We use Key_p shown in Table II as the clues for Percentage category. Besides, some keywords help classify the "relative" and "absolute" subcategories. Writers may use Key_r in Table II to describe the numerals of the "relative" subcategory. If a numeral satisfies with Key_p, but is not described by Key_r, it will be classified as "absolute".

#### 2) Features for Option

Writers usually describe their options with the following pattern:

"maturity date" + "exercise price" + call(put)

Not all descriptions for Option category follow the above pattern, thus we check if there exists "call" or "put" after the numeral, and then check if there exists a month keyword listed in Table II, named Key_m, before the numeral. If both conditions are satisfied, we take this numeral as "maturity date" subcategory. On the other hand, if there does not exist a Key_m, we hold this numeral as "exercise price".

#### 3) Features for Indicator

We consult the indicator name list  to capture a numeral in Indicator category. All indicators in the name list are named Key_i. There are some widely used indicators for investors shown in Table II. If any one of terms in Key_i follows a numeral, then we say that the numeral belongs to the Indicator category.

#### 4) Features for Temporal

We use Key_d and Key_m in Table II as the clues for the subcategory "date", and Key_t is used for "time". If a numeral is followed by any of these keywords, it will be classified into the corresponding subcategory based on the category of the keyword. Furthermore, writers may use some patterns to present a numeral in the "date" subcategory such as "DD/MM/YYYY" and "DD-MM-YY". Thus, we also detect these patterns in tweets and identify the numerals directly. In addition, there are two frequent representations for quarter and half of a year. "Q1", "Q2", "Q3" and "Q4" stand for the first to the fourth quarters. "H1" and "H2" represent the first half and the second half of a year.

#### 5) Features for Quantity

The following pattern is used for the Quantity category:

"Quantity" + Noun

Part-of-speech tagging is performed to check if there is a noun after a numeral. If the numeral and its following term fit this pattern, the numeral will be classified into Quantity category.

### B. Word Vector

Based on 184,050 tweets with 105,255 unique tokens from StockTwits. Skip-gram [14] is used to train word vectors for the financial domain. The dimension is set to 250. In order to reduce the sparseness of numeral patterns in the word embedding, all digits (0 to 9) in the StockTwits corpus are replaced with "D". For example, 8/17 in (T8) will be converted to D/DD, and 5,838 in (T10) will be converted to D,DDD. In these two instances, the pattern D/DD is more likely to be annotated as Temporal than the pattern D,DDD.

### C. *Support Vector Mechine (SVM)*

The SVM model [15] is considered as our baseline classifier. A target numeral in a tweet is represented as two parts. The first part describes the tweet itself, and the second part describes the contextual clues near the numeral according to the features. Here a tweet is encoded as a 4,824-dimensional binary vector, where 4,824 is the vocabulary size of our dataset. The binary value in each dimension indicates if a word appears in the tweet. Finally, we concatenate the 4,824-dimensional vector with the 8-dimensional feature vector to represent a target numeral. Target numerals along with their annotated (sub)categories are used to train SVM.

### D. *Convolutional Neural Network (CNN)*

CNN is one of the popular neural network (NN) models for sentence classification [16]. We encode a target numeral in a tweet with character- and word-based schemes with the matrix composed of three parts as shown in Fig. 1(a).

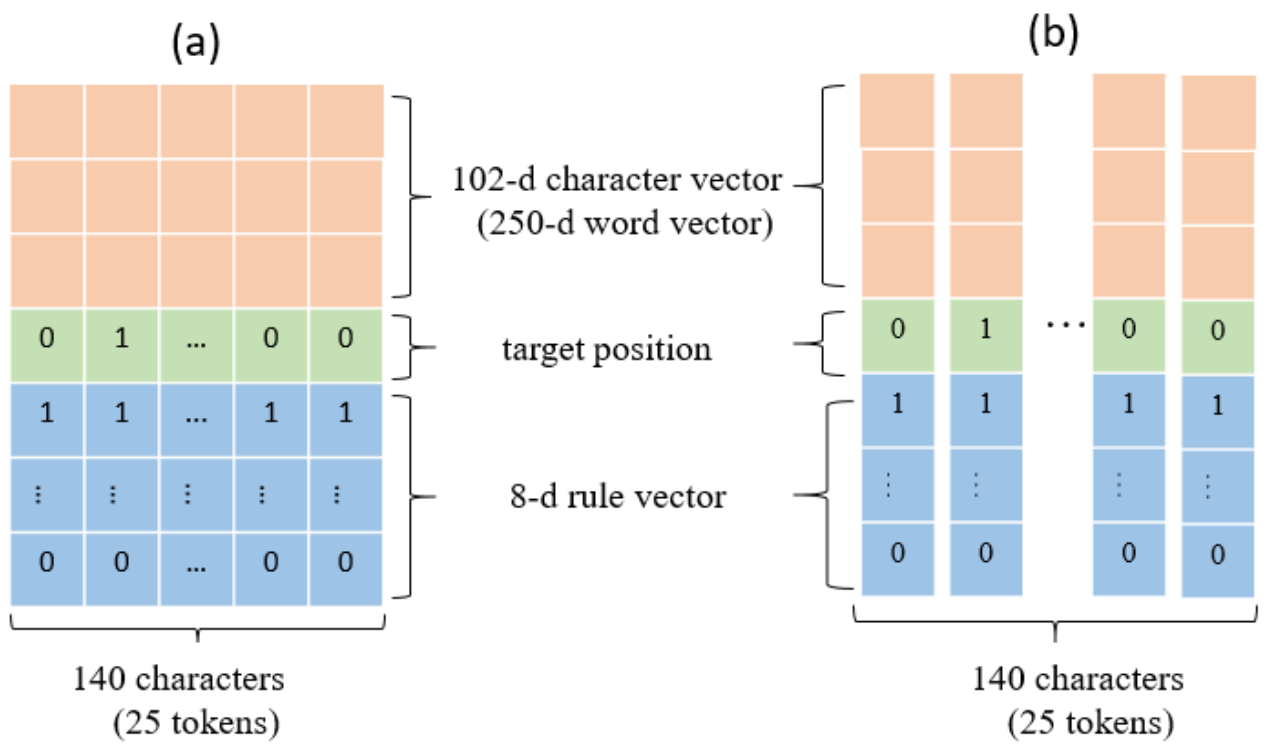

Fig. 1. Representation of a target numeral in a tweet for character-based (word-based) (a) CNN and (b) RNN.

The upper part (in pink), which represents text of a tweet, is an ordered sequence of characters (words) encoded as vectors. In the character-based scheme, a character is represented as a 102-dimensional one-hot vector. The following 102 characters, including basic characters such as digits and alphabets, layout symbols like space, newlines, and tabs, currency symbols (e.g., ¢ and £) and trade-mark symbol (™), constitute our character set.

0123456789
ABCDEFGHIJKLMNOPQRSTUVWXYZ
abcdefghijklmnopqrstuvwxyz
|~™¡¢£¦¨á !"#$%&'( )*+,-./[ ]^_:;<=>?@. \t \n

We set the maximal length of a character sequence to be 140 due to the nature of a tweet. Note that, the data was collected before Twitter expanded the limitation to 280. Zero padding is adopted for tweets less than 140 characters. In the word-based scheme, a word is represented as a 250-dimensional vector pre-trained vector. The longest tweet in our dataset contains 25 tokens, including words, numerals and punctuations. A tweet is encoded as a 25-word sequence with zero padding.

The middle part (in green) specifies the position(s) of the target numeral (there might be multiple contiguous positions in the character-based scheme). The bottom part (in blue) denotes the context of the target numeral in terms of features. In current implementation, we duplicate the 8-dimensional vector 140 (25) times in the character (word)-based scheme.

The reasons for exploring these two schemes are listed below:

**(1) Character vector** can help overcome the problem of informal word forms such as abbreviations in tweets. For instance, in the tweets (T12) and (T13) shown below, we expect that NN models can capture common parts of "res" and "Resistance". Both words are the hints for the "support or resistance" subcategory.

**(2) Word vectors**, which are shown to capture the semantic and syntactic patterns, are widely used in various tasks.

**(T12)** *$ATHX ... very weak res. at 2.51 ............*
**(T13)** *$AMZN new HOD with conviction keeping $570 on watch for Resistance.*

The structure of the proposed CNN model consists of one convolutional layer, one max pooling layer, one densely-connected layer with 64 hidden dimensions, one dropout layer with 0.5 dropout rate, one rectified linear unit (ReLU) layer, and the softmax output layer.

### E. *Recurrent Neural Network (RNN)*

We employ Long Short-Term Memory (LSTM) [17] and Bi-directional LSTM (Bi-LSTM) [18]. The target numeral in a tweet is encoded with character- and word-based schemes as Fig. 1(b). It is regarded as a sequence of vectors. Each vector contains three parts, including representation of the tweet, indication of target position, and 8 features for target numeral. These three parts are interpreted the same as the descriptions for CNN model.

The structure of the proposed RNN models consists of one LSTM (Bi-LSTM) layer with 64 hidden dimensions, one densely-connected layer with 64 hidden dimensions, one dropout layer with 0.5 dropout rate, one ReLU layer, and a softmax output layer.

## VI. EXPERIMENTS

### A. *Preprocessing Procedure*

A financial tweet may be composed of words, cashtags, user id, numbers, URL, hashtags and emojis. In order to reduce the noise, firstly, we replace user ids, cashtags and URLs by "ID", "TICKER" and "URL". Secondly, each digit of the numerals is transformed into "D". Thirdly, we remove emojis, because emojis could present the market sentiment of investors, but could not show the fine-grained opinion defined in our taxonomy. Finally, all remaining tokens are transformed into lowercase.

### B. *Experimental Setup*

Two classification tasks are conducted in the experiments.

**Task 1**: Classify a numeral into 7 categories, i.e., Monetary, Percentage, Option, Indicator, Temporal, Quantity and Product/Version Number.

**Task 2**: Extend the classification task to the subcategory level, and classify numerals into 17 classes, including Indicator, Quantity, Product/Version Number, and all subcategories shown in Table I.

We calculate the micro- and macro-averaged F-scores to evaluate the overall performances of Task 1 and Task 2, and 10-fold cross validation is used. In training process, we split 10% training data into validation set, and use cross entropy loss function. We adopt Adam algorithm [19] to optimize parameters. Early stopping is triggered after 20 trial epochs.

TABLE III. EXPERIMENTAL RESULTS: F-SCORE OF BOTH TASKS (%); *: 95% CONFIDENCE LEVEL

| | | Task1 | | Task2 | |
|---|---|---|---|---|---|
| **Encode** | **Model** | **Micro** | **Macro** | **Micro** | **Macro** |
| | GM | 45.46 | 9.03 | 12.20 | 1.35 |
| | SVM | 59.17 | 42.93 | 13.82 | 3.38 |
| char-based | CNN | 65.85* | 45.33 | **34.99*** | **25.67*** |
| | LSTM | 53.26 | 34.80 | 25.77 | 19.05* |
| | Bi-LSTM | 54.04 | 32.35 | 25.90* | 17.46* |
| word-based | CNN | **67.61*** | **48.34*** | 32.35* | 22.12* |
| | LSTM | 55.05 | 40.65 | 30.61* | 21.39* |
| | Bi-LSTM | 56.96 | 40.98 | 29.59* | 23.36* |

## C. *Experimental Results*

Table III shows the F-scores (%) of different models, and the significance test shows that the improvement is significant with 95% confidence level compared with the SVM model. Guessing majority (GM) approach, which regards the majority class as the prediction result, is the trivial baseline. It only gets macro-averaged F-scores of 9.03% and 1.35% in tasks 1 and 2, respectively. SVM is a strong baseline. It achieves macro-averaged F-scores of 42.93% and 3.38% in the both tasks. The best model in Task 1 is the word-based CNN model. All RNN models do not reach the performances of the SVM model in Task 1. The character-based CNN model performs the best in terms of micro- and macro-averaged F-scores in Task 2. All CNN and RNN models significantly outperform the SVM model in this task. The results of RNN model shows that word-based RNN models are better than character-based RNN models. But RNN models still could not surpass the CNN models when using either encoding methods.

In Task 1, word-based CNN model is only insignificantly better than character-based CNN model. On the other hand, in Task 2, character-based CNN model is significantly better than word-based CNN model with 95% confidence level on the macro-averaged F-score. It shows that character-based CNN model may be the proper model to deal with the work of numeral understanding presented in this paper. One of the possible reasons is that social media posts may be written in informal writing style. With the character-based scheme, the CNN model can capture the patterns of informal writing.

According to the experimental results, the word-based scheme is much more suitable for RNN models. It shows that RNN models may be better in capturing the context information (using word-based scheme) than in capturing the pattern of informal writing (using character-based scheme).

We also experiment with a sequence labeling model that achieves the state-of-the-art performance in NER [20]. On our dataset, the model based on bidirectional LSTM and conditional random field achieves micro-averaged F1-scores of 30.06% and 26.96% in tasks 1 and 2, respectively.

# VII. DISCUSSION

## A. *Analysis of the Models*

Firstly, we do further experiments to deal with the binary problem: whether a target numeral belongs to a certain category or not? Because word-based CNN performs the best in Task 1, we use word-based models to do this experiment.

Fig. 2 shows the results. In the Monetary, Percentage, Option and Indicator categories, word-based CNN models outperform other RNN models. Word-based Bi-LSTM model is good at classifying Temporal and Quantity. All models could not perform well in the Product/ Version Number category. It shows that it is still challenging for the models to capture the product names, e.g., Tesla Model 3 and iPhone 7, using context information without world knowledge. The tailor-made NER model for Product/ Version Number category can be developed in the future.

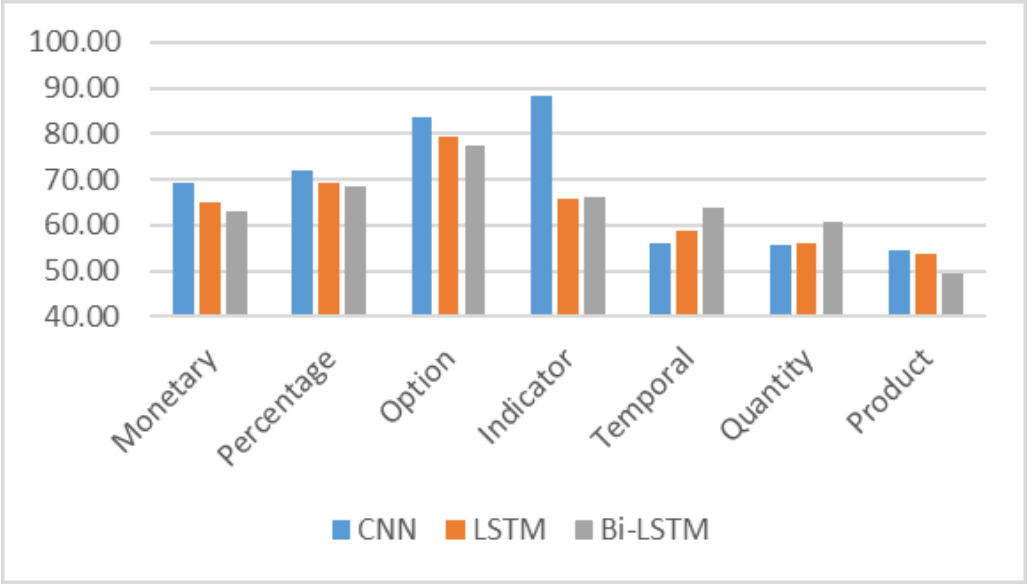

Fig. 2. Macro-averaged F-scroe of 7 categories. (%)

Secondly, we train two-stage models. That is, we use the first classifier to deal with Task 1 and then use the classification result to select a classifier for Task 2. For example, if a numeral is predicted as Monetary by the first classifier, the second classifier will be the classifier trained for the subcategories of the Monetary category. Table IV shows the experimental results. Character-based CNN model still performs the best in these experiments, but none of the two-stage model is better than the character-based CNN model in Table III.

TABLE IV. EXPERIMENTAL RESULTS OF TWO-STAGE MODELS. (%)

| Encode | Model | Micro-averaged F-score (%) | Macro-averaged F-score (%) |
|---|---|---|---|
| char-char | CNN | **34.60** | **24.61** |
| | LSTM | 24.03 | 16.99 |
| | Bi-LS | 26.63 | 17.57 |
| word-word | CNN | 31.78 | 23.17 |
| | LSTM | 27.24 | 18.88 |
| | Bi-LS | 29.79 | 21.51 |
| word-char (Hybrid) | CNN | 32.22 | 24.31 |
| | LSTM | 25.55 | 18.40 |
| | Bi-LS | 28.88 | 19.84 |

Thirdly, because the word-based models outperform the character-based models in Task 1, we further experiment with a hybrid two-stage model. The word-based models are adopted in the first stage to classify a numeral into 7 categories. Then, the character-based models are used to do the fine-grained classification for each category in the second stage. The results are also shown in Table IV. The macro-averaged F-score of the hybrid CNN model is better than that of the one-stage word-based CNN model for Task 2.

We use F1-measure to evaluate the performance of each proposed feature, and show the results of the single-feature-based models in Table V. Percentage, Option and Temporal get higher F1-measure. It validates that the proposed features are useful for the classification task. Furthermore, without the features information, the micro- and macro-averaged F-score of the character-based CNN-model are 8.75% and 5.26% worse than those with features.

TABLE V. PERFORMANCE OF INDIVIDUAL FEATURES. (%)

| Feature | F1 | Feature | F1 |
|---|---|---|---|
| Percentage | 78.19 | Indicator | 42.22 |
| relative | 64.47 | Temporal | 83.18 |
| absolute | 47.56 | date | 52.55 |
| Option | 73.08 | time | 75.00 |
| exercise price | 65.91 | Quantity | 25.84 |
| maturity date | 50.00 | | |

### B. *Error Analysis*

Table VI shows the confusion matrix of Task1. Most errors occur among the Monetary category and other categories. For example, it is very challenging for the model to classify the numerals of Option and Monetary. In the tweet (T17) shown below, 240 is the "exercise price" of Option, but is predicted as Monetary. Actually, it is right, but not precise.

As the performance of human, most of prediction errors of model happen among "quote" and other subcategories. That shows the challenge of understanding the numeral in Monetary category for the machine.

**(T17)** *Sold 1/2 position @$5.70 +$.80 @rsblades Long $AMZN Oct $240 Calls @ $4.90*

TABLE VI. CONFUSION MATRIX OF TASK 1.

| Truth \ Prediction | Temp. | Mone. | Perc. | Option | Indic. | Quan. | Prod. |
|---|---|---|---|---|---|---|---|
| Temporal | 154 | 72 | 5 | 2 | 2 | 1 | 0 |
| Monetary | 51 | 517 | 11 | 6 | 7 | 17 | 0 |
| Percentage | 16 | 91 | 166 | 2 | 3 | 0 | 0 |
| Option | 5 | 8 | 0 | 37 | 2 | 1 | 0 |
| Indicator | 12 | 20 | 4 | 0 | 11 | 1 | 0 |
| Quantity | 17 | 72 | 1 | 0 | 6 | 4 | 0 |
| Product | 2 | 14 | 1 | 0 | 0 | 0 | 0 |

## VIII. CROWD-BASED TRADING MODEL

This section shows an application of numeric opinion mining. We propose a trading model based on numeric opinions made by the crowd. The result of backtest shows the effectiveness of our method on numeral understanding in financial social media data. The comparison of the performances of the crowd and analysts is further analyzed.

### A. *Forecast Price Extraction*

Our trading model relies on the forecast price, which summarizes the analysis results of each investor, in Monetary category. Thus, we sort out the "forecast" price for certain financial instrument of individual investors by extracting the target numerals belong to "forecast" subcategory. Our best CNN model is employed.

### B. *Crowd Opinion vs. Analysts' Opinion*

Some previous works [21, 22] had tried to evaluate the forecast of the professional analysts. However, none of the previous works can evaluate and analyze the opinion of individual investors.

In order to investigate the performance of crowd opinion and analysts' opinion in depth, we extract the forecast price from the tweets mentioned the constituent stocks of Dow Jones Index. We get 91 forecast prices for different stocks in different months. Table VII compares the forecasting performances of the crowd and analysts, where the forecast prices of analysts collected from Bloomberg Terminal are adopted. The average difference between analysts' forecast price and the close price is 6.75%, and that of individual investors is 13.17%. This result shows that the forecasts made by individual investors may have a tendency toward progressive. As a result, individual investors take longer time to achieve the forecast price than professional analysts, and have lower achieving rate.

TABLE VII. CROWD FORECAST PIRCE VS. ANSLSTS' FORECAST PRICE.

| | Crowd | Analyst |
|---|---|---|
| Average difference | 13.17% | 6.75% |
| Achieving rate | 67.03% | 74.73% |
| Achieving duration | 3.38 months | 2.46 months |
| Average return | 4.86% | 2.93% |

A total of 50.59% of extracted forecast price of individual investors and analysts are taking different view (bullish/ bearish) toward the same stock. This results show the crowd's opinion can complement the analysis' results from different aspects, which can eke out the missing part of the analysts. As shown in Fig. 3, analysts' forecast price of $GE (General Electric Company) in September, October, and November shows that they expected the price will rise up in the future. In contrast, that of the individual investors indicate they expect the price will fall down. Finally, the price of $GE fell down since October, and reached crowd's different forecast price in following months. Although analysts lower their forecast price every month, the price has not hit their forecast price till now. That shows the effectiveness of numeric opinion mining from individual investors. Furthermore, sometimes the analysts may not release a report for a certain company. In this case, the crowd-based information can be consulted.

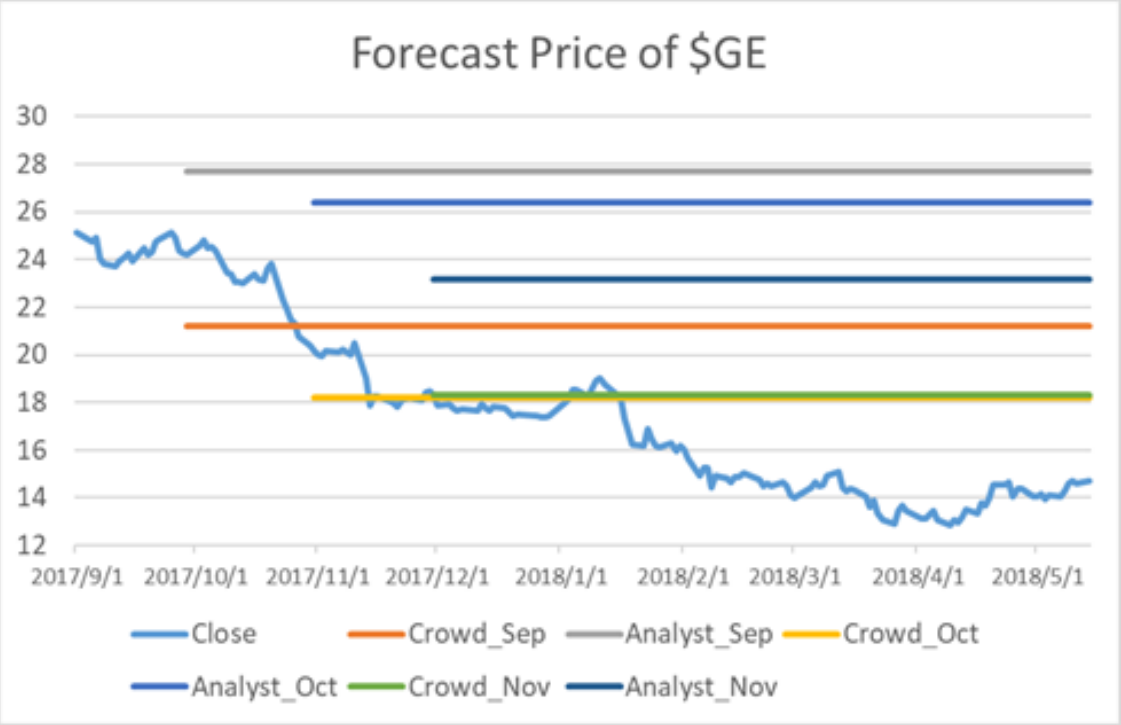

Fig. 3. Crowd's and analysts' forecast price of $GE.

### C. *Trading Strategy*

Based on the extracted "forecast" price, we construct a trading model to test that whether the crowd opinion is the useful signal for earing from the market. We perform the backtest with all constituent stocks, 30 companies, of Dow Jones Index (DJX). Total 26,903 tweets are collected from StockTwits. The duration of social media data is 2017/01/01 to 2017/12/31. The backtest period is 2017/01/01 to 2018/05/14.

Firstly, we check the "forecast" price of each stock at the end of the month. If the forecast price is higher than the close price at the end of the month, we will long the stock. On the other hand, if the forecast price is lower than the close price at the end of the month, we will short the stock.

Secondly, when we are holding the position, we check the close price of each stock in portfolio every day. If the unrealized loss of certain stock reaches 7%, we close whole position of this stock. If the close price reaches the forecast price, we close the position and take the profit.

TABLE VIII. BACKTEST RESULTS OF TRADING STRATEGY.

| | Crowd | Analyst |
|---|---|---|
| Winning ratio | 68.13% | 71.43% |
| Max profit | 52.17% | 17.23% |
| Max drawdown | -11.82% | -14.10% |
| Average profit | 11.08% | 6.42% |
| Average loss | -8.43% | -8.40% |

The backtest results of trading strategy are shown in Table VIII. The trading strategy based on the forecast price made by professional analysts is also provided for comparison. Our trading strategy achieve 68.13% winning ratio, which shows the signals provided by individual investors are trustworthy. With the stop loss condition, our trading strategy control the drawdown within 11.82%. The average return is 4.86%.

With the narrow forecast price, analysts achieve higher winning ratio than individual investors do. However, the overall performance of analysts is worse than individual investors. On the one hand, the average return of analysts is lower than that of individual investors. This result indicates that following the conservative forecast price of analysts may cause investors close their position too early. The comparison of the max profit and average profit also shows the evidence for this phenomenon. On the other hand, from the aspect of the risk investors may take, the average loss of both trading strategies are close, but the max drawdown of the strategy that follows the forecast price of analysts is higher than that of the strategy following the forecast price of individual investors. All evidences show that using the crowd opinion to predict the market movement may better than using the opinion of analysts.

## IX. CONCLUSION

In this paper, we address a new opinion mining challenge to capture the view of the investors on the social media platform by giving a fine-grained taxonomy for numerals in financial tweets. We compare the SVM, CNN and RNN models with different representations of a target numeral. Experimental results show that the word-based CNN model achieves the best performance in coarse-grained classification task, and the character-based CNN model achieves the best performance in fine-grained classification task.

Previous research constructing the trading strategies based on sentiment scores can only get the bullish/ bearish view of the investors, but cannot get the information of the time to close their position. In this paper, based on the forecast price of the individual investors, we provide the trading strategy that has both bullish/bearish information and the price level to close the position. The backtest results of trading strategy indicate that capturing fine-grained crowd opinion is promising. We overcome the challenge of evaluating crowd opinion, and provide the first comparison of the forecast price between individual investors and professional analysts. With our dataset and models, lots of extended application scenarios can be addressed. For example, we can comprehend the "support or resistance" of investors, which implies that we know when the investors will buy/ sell their position. We release the annotated dataset, FinNum, as a resource for research purpose.

## ACKNOWLEDGEMENTS

This research was partially supported by Ministry of Science and Technology, Taiwan, under grants MOST-106-2923-E-002-012-MY3, MOST-107-2634-F-002-011-, and MOST-107-2218-E-009-050-.